\documentclass[sigconf]{acmart-me}

\usepackage{booktabs} 
\usepackage{url}
\usepackage{color}
\usepackage{enumitem}
\usepackage{subcaption}
\usepackage{balance}
\usepackage[show]{inotes}

\setcopyright{rightsretained}

\acmDOI{}

\acmISBN{}

\acmConference[MediaEval'20]{Multimedia Evaluation Workshop}{December 14-15 2020}{Online} 
\acmYear{2020}
\copyrightyear{}

\acmPrice{}

\begin{document}
\title{Flood Detection via Twitter Streams using Textual and Visual Features}

\author{Firoj Alam\textsuperscript{1}, Zohaib Hassan \textsuperscript{2}, Kashif Ahmad \textsuperscript{3}, Asma Gul \textsuperscript{4}, Michael Reiglar \textsuperscript{5}, Nicola Conci \textsuperscript{2}, Ala Al-Fuqaha \textsuperscript{3}}
\affiliation{\textsuperscript{1}Qatar Computing Research Institute, Doha, Qatar,  \textsuperscript{2} University of Trento, Italy \\ \textsuperscript{3} Division of Information and Computing Technology, College of Science and Engineering, Hamad Bin Khalifa University, Qatar Foundation, Doha, Qatar,\textsuperscript{4} Department of Statistics, Shaheed Benazir Bhutto Women University, Peshawar, Pakistan,\textsuperscript{5} SimulaMet, Norway}

%
%
%
%
%

\renewcommand{\shortauthors}{F. Alam et al.}
\renewcommand{\shorttitle}{Flood-related Multimedia}

\begin{abstract}
The paper presents our proposed solutions for the MediaEval 2020 Flood-Related Multimedia Task, which aims to analyze and detect flooding events in multimedia content shared over Twitter. In total, we proposed four different solutions including a multi-modal solution combining textual and visual information for the mandatory run, and three single modal image and text-based solutions as optional runs. In the multi-modal method, we rely on a supervised multimodal bitransformer model that combines textual and visual features in an early fusion, achieving a micro F1-score of .859 on the development data set. For the text-based flood events detection, we use a transformer network (i.e., pretrained Italian BERT model) achieving an F1-score of .853. For image-based solutions, we employed multiple deep models, pre-trained on both, the ImageNet and places data sets, individually and combined in an early fusion achieving F1-scores of .816 and .805 on the development set, respectively.

\end{abstract}

%
%
%
%
%


\maketitle

\section{Introduction}
\label{sec:intro}
Floods are the most frequent and devastating type of natural disaster causing a significant loss in terms of human lives and infrastructure worldwide, every year. According to a recent report \cite{WHO_Floods}, around 80-90\% of natural disasters worldwide over the last decade are caused by floods, and more than two billion people worldwide were affected between 1998-2017. The damage of flood can be significantly mitigated if timely and accurate information about the location, scale, and most affected areas is available \cite{ahmad2018social,Said2019}. However, several challenges, such as the availability of reporters and other resources, etc., are associated with the gathering of such information during floods \cite{Said2019}. On the other hand, social media has been proved very effective in information dissemination in such events \cite{Said2019,ahmad2019automatic,bischke2016contextual,alam2020descriptive,ofli2020analysis}.   
 
Similar to previous years of the challenge \cite{bischke2017multimedia,bischke2018multimediasatellite,bischke2019mmsat}, the MediaEval 2020 flood-related multimedia task \cite{andreadis2020floodmultimedia} aims to analyze tweets from Twitter for flood events detection. The participants were provided a collection of tweets with associated images, and were asked to propose a framework able to automatically identify floods related tweets relevant to a particular area. This paper provides the details of the methods proposed by team HBKU\_UNITN\_SIMULA for the task. In total, we proposed four different solutions including a multi-modal one for the mandatory run, a textual information based solution, and a couple of image-based solutions for flood events detection in Twitter images. 

\section{Proposed Approaches}
Transfer learning has become mainstream in computer vision and natural language processing (NLP). For example, in computer vision models (e.g., VGG16, ResNet18) trained using ImageNet \cite{krizhevsky2012imagenet} or  Places Database \cite{ zhou2014learning} have been used as pre-trained models to initialize networks for fine-tunning the task-specific models. For NLP, word-embedding \cite{mikolov2013distributed}, sentence-embedding \cite{conneau2017supervised}, and recent BERT \cite{devlin2018bert} based models have shown significant progresses in downstream tasks. 
For this study, we used deep CNN models for image classification and a transformer model for text classification and finally combine them to design a multimodel network. 
Prior work with similar approaches in this direction include \cite{kiela2019supervised,ofli2020analysis,abavisani2020multimodal,Agarwal_Leekha_Sawhney_Shah_2020}. 
The study in \cite{kiela2019supervised} used a combination of a deep CNN for image and a transformer model for text in a similar manner as proposed in this work. 
For the disaster response task, in \cite{ofli2020analysis} the authors used a deep CNN (VGG16) for the image, a CNN with static embedding for text, and finally combine them the shared representation before a softmax layer for classification. 
In \cite{abavisani2020multimodal}, the authors propose a cross-attention module for multimodal fusion, and the study in \cite{Agarwal_Leekha_Sawhney_Shah_2020} proposes different fusion approaches for combining disaster-related tweet classification tasks. 
Our work is in line with the study by \cite{kiela2019supervised}, however, our work is different in how we use different pre-trained models (i.e., models trained with ImageNet and Places database). In the next section we discuss the detail of the models used in this study. 

\subsection{Text-based Model (Run 2)}
\paragraph{Pre-processing} 
For the text-based model, we first pre-process the tweet texts as they are noisy, and consists of many symbols, emoticons, URLs, usernames, and invisible characters. Prior studies like \cite{ofli2020analysis} show that filtering and cleaning the tweets before training a classifier helps significantly. We pre-process the tweet texts before the classification experiments. The preprocessing include removal of invisible characters, URLs, and hashtag signs. 
\paragraph{Transformer model} 
The pre-processed texts are then feed into a transformer network by adding a task-specific layer on top of the network. We use model specific tokenizer, which is a part of the transformer model. Currently, the pre-trained transformer models are available for monolingual and multilingual settings. Since the tweets are in the Italian language, and for Italian a monolingual model exists, namely Italian BERT\footnote{\url{https://huggingface.co/dbmdz/bert-base-italian-uncased}}\footnote{Note that the model is trained on Wikipedia dump and various texts from the OPUS corpora.}, we used it for our experiments.

For the training, we used the Transformer Toolkit~\cite{Wolf2019HuggingFacesTS}. We fine-tune the model using a learning rate of $1e-5$ for ten epochs \cite{devlin2018bert}. The training of the pre-trained models has some instability as reported in~\cite{devlin2018bert}, therefore, we run each experiment ten times using different seeds and select the model that performs the best on the development set. Finally, we evaluate the selected model on the test set. 

\subsection{Image-based Model (Run 3 and 4)}
\label{ssec:image-based}
For the flood image detection we employed two different methods. In the first method, we fine-tuned an existing model, namely VggNet16 \cite{simonyan2014very}, pre-trained on the Places dataset \cite{zhou2014learning}. In the second method, we jointly utilized the models pre-trained on the ImageNet \cite{deng2009imagenet} and the Places dataset. The basic motivation for the joint use of the models comes from our previous experience on similar tasks \cite{ahmad2019automatic,ahmad2017convolutional}, where the fusion of object and scene-level information extracted with the models pre-trained on ImageNet and Places dataset, respectively, have been proven very effective in classification of disaster-related images. 

The class distribution of the dataset for the challenge this year is very imbalanced. Therefore, we used an oversampling technique to balance the distribution of the class labels in the training. This of course comes with the risk that the test data might not be imbalanced which would most probably reduce the performance. We used the Synthetic Minority Oversampling Technique (SMOTE) \cite{chawla2002smote} to up-sample the minority class. We used the \textit{imblearn} implementation~\cite{JMLR:v18:16-365} for our experiments. In fact, the number of samples in the minority class have been increased by a factor of three.

\subsection{Multimodal Model (Run 1)}
\label{ssec:multimodal}
The multimodal network consists of a text and an image network combined to form a shared representation before the classification layer. The text network consists of BERT \cite{devlin2018bert} and the image network consists of ResNet152~\cite{he2016deep}. We used ResNet152 in multimodal network as it was shown to work well in a previous study \cite{kiela2019supervised}.
The input to the whole network is pre-processed text and extracted features for the images. The object and scene-level features are extracted through VGGNet16 pre-trained on ImageNet and Places datasets. 
During the training the model jointly learn the image embeddings and token embedding spaces of BERT. 
We use the Adam optimizer with a minibatch size of $32$ for training the model. 

\section{Results and Analysis}
\label{sec:results}
In total, we submitted four runs. Our first run is based on the multimodal framework where textual and visual features are combined using early fusion. 
Our second run is based on an Italian version of the BERT model for text analysis. The third run is based on VGGNet-19 pre-trained on Places datasets, which is fine-tuned on the floods-related images. In run four, two versions of VGGNet-16, one pre-trained on ImageNet and the other pre-trained the places dataset, are combined in an early fusion manner by concatenating the features obtained from the last fully connected layer. 

Table \ref{tab:experimental_results} provides the experimental results of our proposed solutions for the task on the development set. Overall better results are obtained with run 1, which shows the advantage of a multi-modal solution over the single modality in the task. On the other hand, the lowest F1-score is obtained with fusion of object and scene-level features. However, interesting is that the results obtained with the individual model (VGGNet16) pre-trained on the places dataset has outperformed the method combining the object and scene-level features. In order to 
investigate the potential causes of the reduction in the performance of the fusion framework, we also analyzed the performance of VGGNet16 pre-trained on ImageNet where an F1-score of .804 is obtained. The lower performance of the model when pre-trained on ImageNet compared to the Places data indicates that the scene-level features are more important for the task. It is to be noted that the models pre-trained on ImageNet extract object-level while the ones pre-trained on the Places datasets correspond to scene-level information. Due to the imbalanced in the data set we decided not to discuss the test data set results before the test data was publicly released and we are able to perform a more detailed analysis.

\begin{table}[]
\centering
\caption{Experimental results of the proposed methods.}
\label{tab:experimental_results}
\begin{tabular}{|c|c|c|}
\hline
\textbf{Run \#} & \textbf{Type of Features} & \textbf{Micro F1-Score} \\ \hline
 1 & Textual + Visual & 0.859 \\ \hline
2 & Textual  & 0.853 \\ \hline
 3 & Visual & 0.816 \\ \hline
  4 & Visual &0.805  \\ \hline
\end{tabular}
\end{table}
\section{Conclusions and Future Work}
The task aims at the multimodal analysis of floods on Twitter. The participants were provided with a collection of tweets containing text and associated images and were asked to propose a multimodal framework able to automatically determine whether a tweet represents a flood-related event relevant to a specific area or not. 
We proposed four different solutions to the task including a multimodal, a textual, and a couple of image-based solutions. Overall, better results are observed for the multimodal approach indicating the advantage of the joint use of textual and visual information. As far as the evaluation of textual and visual information is concerned, significantly better results are obtained with textual features compared to visual information. Moreover, we also observed that scene-level information is more critical for the task compared to object-level features extracted with models pre-trained on ImageNet.

We believe there is still room for improvement in the multimodal, textual, and image-based solutions. In the future, we aim to explore the task further by introducing more sophisticated methods to jointly combine textual and visual information in a better way. We also plan to perform a more detailed analysis of the test data once publicly released. 
\balance

\bibliographystyle{ACM-Reference-Format}
\def\bibfont{\small} 
\bibliography{sigproc} 


\begin{thebibliography}{00}


\ifx \showCODEN    \undefined \def \showCODEN     #1{\unskip}     \fi
\ifx \showDOI      \undefined \def \showDOI       #1{#1}\fi
\ifx \showISBNx    \undefined \def \showISBNx     #1{\unskip}     \fi
\ifx \showISBNxiii \undefined \def \showISBNxiii  #1{\unskip}     \fi
\ifx \showISSN     \undefined \def \showISSN      #1{\unskip}     \fi
\ifx \showLCCN     \undefined \def \showLCCN      #1{\unskip}     \fi
\ifx \shownote     \undefined \def \shownote      #1{#1}          \fi
\ifx \showarticletitle \undefined \def \showarticletitle #1{#1}   \fi
\ifx \showURL      \undefined \def \showURL       {\relax}        \fi
\providecommand\bibfield[2]{#2}
\providecommand\bibinfo[2]{#2}
\providecommand\natexlab[1]{#1}
\providecommand\showeprint[2][]{arXiv:#2}

\bibitem[\protect\citeauthoryear{??}{WHO}{}]%
        {WHO_Floods}
\bibinfo{title}{WHO Report on Floods}.
\newblock \bibinfo{howpublished}{\url{https://tinyurl.com/y5qlldbn}}.
  (\bibinfo{year}{????}).
\newblock
\newblock
\shownote{Accessed: 2020-11-17.}


\bibitem[\protect\citeauthoryear{Abavisani, Wu, Hu, Tetreault, and
  Jaimes}{Abavisani et~al\mbox{.}}{2020}]%
        {abavisani2020multimodal}
\bibfield{author}{\bibinfo{person}{Mahdi Abavisani}, \bibinfo{person}{Liwei
  Wu}, \bibinfo{person}{Shengli Hu}, \bibinfo{person}{Joel Tetreault}, {and}
  \bibinfo{person}{Alejandro Jaimes}.} \bibinfo{year}{2020}\natexlab{}.
\newblock \showarticletitle{Multimodal Categorization of Crisis Events in
  Social Media}. In \bibinfo{booktitle}{{\em Proceedings of the IEEE/CVF
  Conference on Computer Vision and Pattern Recognition}}.
  \bibinfo{pages}{14679--14689}.
\newblock


\bibitem[\protect\citeauthoryear{Agarwal, Leekha, Sawhney, and Shah}{Agarwal
  et~al\mbox{.}}{2020}]%
        {Agarwal_Leekha_Sawhney_Shah_2020}
\bibfield{author}{\bibinfo{person}{Mansi Agarwal}, \bibinfo{person}{Maitree
  Leekha}, \bibinfo{person}{Ramit Sawhney}, {and} \bibinfo{person}{Rajiv~Ratn
  Shah}.} \bibinfo{year}{2020}\natexlab{}.
\newblock \showarticletitle{Crisis-DIAS: Towards Multimodal Damage Analysis -
  Deployment, Challenges and Assessment}.
\newblock \bibinfo{journal}{{\em Proceedings of the AAAI Conference on
  Artificial Intelligence\/}} \bibinfo{volume}{34}, \bibinfo{number}{01}
  (\bibinfo{date}{Apr.} \bibinfo{year}{2020}), \bibinfo{pages}{346--353}.
\newblock
\showDOI{%
\url{https://doi.org/10.1609/aaai.v34i01.5369}}


\bibitem[\protect\citeauthoryear{Ahmad, Pogorelov, Riegler, Conci, and
  Halvorsen}{Ahmad et~al\mbox{.}}{2018}]%
        {ahmad2018social}
\bibfield{author}{\bibinfo{person}{Kashif Ahmad}, \bibinfo{person}{Konstantin
  Pogorelov}, \bibinfo{person}{Michael Riegler}, \bibinfo{person}{Nicola
  Conci}, {and} \bibinfo{person}{P{\aa}l Halvorsen}.}
  \bibinfo{year}{2018}\natexlab{}.
\newblock \showarticletitle{Social media and satellites}.
\newblock \bibinfo{journal}{{\em Multimedia Tools and Applications\/}}
  (\bibinfo{year}{2018}), \bibinfo{pages}{1--39}.
\newblock


\bibitem[\protect\citeauthoryear{Ahmad, Pogorelov, Riegler, Ostroukhova,
  Halvorsen, Conci, and Dahyot}{Ahmad et~al\mbox{.}}{2019}]%
        {ahmad2019automatic}
\bibfield{author}{\bibinfo{person}{Kashif Ahmad}, \bibinfo{person}{Konstantin
  Pogorelov}, \bibinfo{person}{Michael Riegler}, \bibinfo{person}{Olga
  Ostroukhova}, \bibinfo{person}{P{\aa}l Halvorsen}, \bibinfo{person}{Nicola
  Conci}, {and} \bibinfo{person}{Rozenn Dahyot}.}
  \bibinfo{year}{2019}\natexlab{}.
\newblock \showarticletitle{Automatic detection of passable roads after floods
  in remote sensed and social media data}.
\newblock \bibinfo{journal}{{\em Signal Processing: Image Communication\/}}
  \bibinfo{volume}{74} (\bibinfo{year}{2019}), \bibinfo{pages}{110--118}.
\newblock


\bibitem[\protect\citeauthoryear{Ahmad, Ahmad, Ahmad, and Conci}{Ahmad
  et~al\mbox{.}}{2017}]%
        {ahmad2017convolutional}
\bibfield{author}{\bibinfo{person}{Sheharyar Ahmad}, \bibinfo{person}{Kashif
  Ahmad}, \bibinfo{person}{Nasir Ahmad}, {and} \bibinfo{person}{Nicola Conci}.}
  \bibinfo{year}{2017}\natexlab{}.
\newblock \showarticletitle{Convolutional neural networks for disaster images
  retrieval}. In \bibinfo{booktitle}{{\em Proceedings of the MediaEval 2017
  Workshop (Sept. 13--15, 2017). Dublin, Ireland}}.
\newblock


\bibitem[\protect\citeauthoryear{Alam, Ofli, and Imran}{Alam
  et~al\mbox{.}}{2020}]%
        {alam2020descriptive}
\bibfield{author}{\bibinfo{person}{Firoj Alam}, \bibinfo{person}{Ferda Ofli},
  {and} \bibinfo{person}{Muhammad Imran}.} \bibinfo{year}{2020}\natexlab{}.
\newblock \showarticletitle{Descriptive and visual summaries of disaster events
  using artificial intelligence techniques: case studies of Hurricanes Harvey,
  Irma, and Maria}.
\newblock \bibinfo{journal}{{\em Behaviour \& Information Technology\/}}
  \bibinfo{volume}{39}, \bibinfo{number}{3} (\bibinfo{year}{2020}),
  \bibinfo{pages}{288--318}.
\newblock


\bibitem[\protect\citeauthoryear{Andreadis, Gialampoukidis, Karakostas,
  Vrochidis, Kompatsiaris, Fiorin, Norbiato, and Ferri}{Andreadis
  et~al\mbox{.}}{2020}]%
        {andreadis2020floodmultimedia}
\bibfield{author}{\bibinfo{person}{Stelios Andreadis}, \bibinfo{person}{Ilias
  Gialampoukidis}, \bibinfo{person}{Anastasios Karakostas},
  \bibinfo{person}{Stefanos Vrochidis}, \bibinfo{person}{Ioannis Kompatsiaris},
  \bibinfo{person}{Roberto Fiorin}, \bibinfo{person}{Daniele Norbiato}, {and}
  \bibinfo{person}{Michele Ferri}.} \bibinfo{year}{2020}\natexlab{}.
\newblock \showarticletitle{The Flood-related Multimedia Task at MediaEval
  2020}.
\newblock  (\bibinfo{year}{2020}).
\newblock


\bibitem[\protect\citeauthoryear{Bischke, Borth, Schulze, and Dengel}{Bischke
  et~al\mbox{.}}{2016}]%
        {bischke2016contextual}
\bibfield{author}{\bibinfo{person}{Benjamin Bischke}, \bibinfo{person}{Damian
  Borth}, \bibinfo{person}{Christian Schulze}, {and} \bibinfo{person}{Andreas
  Dengel}.} \bibinfo{year}{2016}\natexlab{}.
\newblock \showarticletitle{Contextual enrichment of remote-sensed events with
  social media streams}. In \bibinfo{booktitle}{{\em Proceedings of the 24th
  ACM international conference on Multimedia}}. ACM,
  \bibinfo{pages}{1077--1081}.
\newblock


\bibitem[\protect\citeauthoryear{Bischke, Helber, Basar, Brugman, Zhao, and
  Pogorelov}{Bischke et~al\mbox{.}}{}]%
        {bischke2019mmsat}
\bibfield{author}{\bibinfo{person}{Benjamin Bischke}, \bibinfo{person}{Patrick
  Helber}, \bibinfo{person}{Erkan Basar}, \bibinfo{person}{Simon Brugman},
  \bibinfo{person}{Zhengyu Zhao}, {and} \bibinfo{person}{Konstantin
  Pogorelov}.}
\newblock \showarticletitle{The Multimedia Satellite Task at MediaEval 2019:
  Flood Severity Estimation}. In \bibinfo{booktitle}{{\em Proc. of the
  MediaEval 2019 Workshop}} (Oct. 27-29, 2019). \bibinfo{address}{Sophia
  Antipolis, France}.
\newblock


\bibitem[\protect\citeauthoryear{Bischke, Helber, Schulze, Venkat, Dengel, and
  Borth}{Bischke et~al\mbox{.}}{2017}]%
        {bischke2017multimedia}
\bibfield{author}{\bibinfo{person}{Benjamin Bischke}, \bibinfo{person}{Patrick
  Helber}, \bibinfo{person}{Christian Schulze}, \bibinfo{person}{Srinivasan
  Venkat}, \bibinfo{person}{Andreas Dengel}, {and} \bibinfo{person}{Damian
  Borth}.} \bibinfo{year}{2017}\natexlab{}.
\newblock \showarticletitle{The Multimedia Satellite Task at MediaEval 2017:
  Emergence Response for Flooding Events}. In \bibinfo{booktitle}{{\em
  Proceedings of the MediaEval 2017 Workshop (Sept. 13-15, 2017). Dublin,
  Ireland}}.
\newblock


\bibitem[\protect\citeauthoryear{Bischke, Helber, Zhao, de~Bruijn, and
  Borth}{Bischke et~al\mbox{.}}{}]%
        {bischke2018multimediasatellite}
\bibfield{author}{\bibinfo{person}{Benjamin Bischke}, \bibinfo{person}{Patrick
  Helber}, \bibinfo{person}{Zhengyu Zhao}, \bibinfo{person}{Jens de Bruijn},
  {and} \bibinfo{person}{Damian Borth}.}
\newblock \showarticletitle{The Multimedia Satellite Task at MediaEval 2018:
  Emergency Response for Flooding Events}. In \bibinfo{booktitle}{{\em Proc. of
  the MediaEval 2018 Workshop}} (Oct. 29-31, 2018).
  \bibinfo{address}{Sophia-Antipolis, France}.
\newblock


\bibitem[\protect\citeauthoryear{Chawla, Bowyer, Hall, and Kegelmeyer}{Chawla
  et~al\mbox{.}}{2002}]%
        {chawla2002smote}
\bibfield{author}{\bibinfo{person}{Nitesh~V Chawla}, \bibinfo{person}{Kevin~W
  Bowyer}, \bibinfo{person}{Lawrence~O Hall}, {and} \bibinfo{person}{W~Philip
  Kegelmeyer}.} \bibinfo{year}{2002}\natexlab{}.
\newblock \showarticletitle{SMOTE: synthetic minority over-sampling technique}.
\newblock \bibinfo{journal}{{\em Journal of artificial intelligence
  research\/}}  \bibinfo{volume}{16} (\bibinfo{year}{2002}),
  \bibinfo{pages}{321--357}.
\newblock


\bibitem[\protect\citeauthoryear{Conneau, Kiela, Schwenk, Barrault, and
  Bordes}{Conneau et~al\mbox{.}}{2017}]%
        {conneau2017supervised}
\bibfield{author}{\bibinfo{person}{Alexis Conneau}, \bibinfo{person}{Douwe
  Kiela}, \bibinfo{person}{Holger Schwenk}, \bibinfo{person}{Loic Barrault},
  {and} \bibinfo{person}{Antoine Bordes}.} \bibinfo{year}{2017}\natexlab{}.
\newblock \showarticletitle{Supervised learning of universal sentence
  representations from natural language inference data}.
\newblock \bibinfo{journal}{{\em arXiv preprint arXiv:1705.02364\/}}
  (\bibinfo{year}{2017}).
\newblock


\bibitem[\protect\citeauthoryear{Deng, Dong, Socher, Li, Li, and Fei-Fei}{Deng
  et~al\mbox{.}}{2009}]%
        {deng2009imagenet}
\bibfield{author}{\bibinfo{person}{Jia Deng}, \bibinfo{person}{Wei Dong},
  \bibinfo{person}{Richard Socher}, \bibinfo{person}{Li-Jia Li},
  \bibinfo{person}{Kai Li}, {and} \bibinfo{person}{Li Fei-Fei}.}
  \bibinfo{year}{2009}\natexlab{}.
\newblock \showarticletitle{Imagenet: A large-scale hierarchical image
  database}. In \bibinfo{booktitle}{{\em Computer Vision and Pattern
  Recognition, 2009. CVPR 2009. IEEE Conference on}}. Ieee,
  \bibinfo{pages}{248--255}.
\newblock


\bibitem[\protect\citeauthoryear{Devlin, Chang, Lee, and Toutanova}{Devlin
  et~al\mbox{.}}{2018}]%
        {devlin2018bert}
\bibfield{author}{\bibinfo{person}{Jacob Devlin}, \bibinfo{person}{Ming-Wei
  Chang}, \bibinfo{person}{Kenton Lee}, {and} \bibinfo{person}{Kristina
  Toutanova}.} \bibinfo{year}{2018}\natexlab{}.
\newblock \showarticletitle{{BERT:} Pre-training of deep bidirectional
  transformers for language understanding}.
\newblock \bibinfo{journal}{{\em arXiv preprint arXiv:1810.04805\/}}
  (\bibinfo{year}{2018}).
\newblock


\bibitem[\protect\citeauthoryear{He, Zhang, Ren, and Sun}{He
  et~al\mbox{.}}{2016}]%
        {he2016deep}
\bibfield{author}{\bibinfo{person}{Kaiming He}, \bibinfo{person}{Xiangyu
  Zhang}, \bibinfo{person}{Shaoqing Ren}, {and} \bibinfo{person}{Jian Sun}.}
  \bibinfo{year}{2016}\natexlab{}.
\newblock \showarticletitle{Deep residual learning for image recognition}. In
  \bibinfo{booktitle}{{\em Proceedings of the IEEE conference on computer
  vision and pattern recognition}}. \bibinfo{pages}{770--778}.
\newblock


\bibitem[\protect\citeauthoryear{Kiela, Bhooshan, Firooz, and Testuggine}{Kiela
  et~al\mbox{.}}{2019}]%
        {kiela2019supervised}
\bibfield{author}{\bibinfo{person}{Douwe Kiela}, \bibinfo{person}{Suvrat
  Bhooshan}, \bibinfo{person}{Hamed Firooz}, {and} \bibinfo{person}{Davide
  Testuggine}.} \bibinfo{year}{2019}\natexlab{}.
\newblock \showarticletitle{Supervised multimodal bitransformers for
  classifying images and text}.
\newblock \bibinfo{journal}{{\em arXiv preprint arXiv:1909.02950\/}}
  (\bibinfo{year}{2019}).
\newblock


\bibitem[\protect\citeauthoryear{Krizhevsky, Sutskever, and Hinton}{Krizhevsky
  et~al\mbox{.}}{2012}]%
        {krizhevsky2012imagenet}
\bibfield{author}{\bibinfo{person}{Alex Krizhevsky}, \bibinfo{person}{Ilya
  Sutskever}, {and} \bibinfo{person}{Geoffrey~E Hinton}.}
  \bibinfo{year}{2012}\natexlab{}.
\newblock \showarticletitle{Imagenet classification with deep convolutional
  neural networks}. In \bibinfo{booktitle}{{\em Advances in neural information
  processing systems}}. \bibinfo{pages}{1097--1105}.
\newblock


\bibitem[\protect\citeauthoryear{Lema{{\^i}}tre, Nogueira, and
  Aridas}{Lema{{\^i}}tre et~al\mbox{.}}{2017}]%
        {JMLR:v18:16-365}
\bibfield{author}{\bibinfo{person}{Guillaume Lema{{\^i}}tre},
  \bibinfo{person}{Fernando Nogueira}, {and} \bibinfo{person}{Christos~K.
  Aridas}.} \bibinfo{year}{2017}\natexlab{}.
\newblock \showarticletitle{Imbalanced-learn: A Python Toolbox to Tackle the
  Curse of Imbalanced Datasets in Machine Learning}.
\newblock \bibinfo{journal}{{\em Journal of Machine Learning Research\/}}
  \bibinfo{volume}{18}, \bibinfo{number}{17} (\bibinfo{year}{2017}),
  \bibinfo{pages}{1--5}.
\newblock
\showURL{%
\url{http://jmlr.org/papers/v18/16-365}}


\bibitem[\protect\citeauthoryear{Mikolov, Sutskever, Chen, Corrado, and
  Dean}{Mikolov et~al\mbox{.}}{2013}]%
        {mikolov2013distributed}
\bibfield{author}{\bibinfo{person}{Tomas Mikolov}, \bibinfo{person}{Ilya
  Sutskever}, \bibinfo{person}{Kai Chen}, \bibinfo{person}{Greg~S Corrado},
  {and} \bibinfo{person}{Jeff Dean}.} \bibinfo{year}{2013}\natexlab{}.
\newblock \showarticletitle{Distributed representations of words and phrases
  and their compositionality}. In \bibinfo{booktitle}{{\em Advances in neural
  information processing systems}}. \bibinfo{pages}{3111--3119}.
\newblock


\bibitem[\protect\citeauthoryear{Ofli, Alam, and Imran}{Ofli
  et~al\mbox{.}}{2020}]%
        {ofli2020analysis}
\bibfield{author}{\bibinfo{person}{Ferda Ofli}, \bibinfo{person}{Firoj Alam},
  {and} \bibinfo{person}{Muhammad Imran}.} \bibinfo{year}{2020}\natexlab{}.
\newblock \showarticletitle{Analysis of Social Media Data using Multimodal Deep
  Learning for Disaster Response}. In \bibinfo{booktitle}{{\em ISCRAM}}.
\newblock


\bibitem[\protect\citeauthoryear{Said, Ahmad, Riegler, Pogorelov, Hassan,
  Ahmad, and Conci}{Said et~al\mbox{.}}{2019}]%
        {Said2019}
\bibfield{author}{\bibinfo{person}{Naina Said}, \bibinfo{person}{Kashif Ahmad},
  \bibinfo{person}{Michael Riegler}, \bibinfo{person}{Konstantin Pogorelov},
  \bibinfo{person}{Laiq Hassan}, \bibinfo{person}{Nasir Ahmad}, {and}
  \bibinfo{person}{Nicola Conci}.} \bibinfo{year}{2019}\natexlab{}.
\newblock \showarticletitle{Natural disasters detection in social media and
  satellite imagery: a survey}.
\newblock \bibinfo{journal}{{\em Multimedia Tools and Applications\/}}
  (\bibinfo{date}{17 Jul} \bibinfo{year}{2019}).
\newblock
\showISSN{1573-7721}
\showDOI{%
\url{https://doi.org/10.1007/s11042-019-07942-1}}


\bibitem[\protect\citeauthoryear{Simonyan and Zisserman}{Simonyan and
  Zisserman}{2014}]%
        {simonyan2014very}
\bibfield{author}{\bibinfo{person}{Karen Simonyan} {and}
  \bibinfo{person}{Andrew Zisserman}.} \bibinfo{year}{2014}\natexlab{}.
\newblock \showarticletitle{Very deep convolutional networks for large-scale
  image recognition}.
\newblock \bibinfo{journal}{{\em arXiv preprint arXiv:1409.1556\/}}
  (\bibinfo{year}{2014}).
\newblock


\bibitem[\protect\citeauthoryear{Wolf, Debut, Sanh, Chaumond, Delangue, Moi,
  Cistac, Rault, Louf, Funtowicz, and Brew}{Wolf et~al\mbox{.}}{2019}]%
        {Wolf2019HuggingFacesTS}
\bibfield{author}{\bibinfo{person}{Thomas Wolf}, \bibinfo{person}{Lysandre
  Debut}, \bibinfo{person}{Victor Sanh}, \bibinfo{person}{Julien Chaumond},
  \bibinfo{person}{Clement Delangue}, \bibinfo{person}{Anthony Moi},
  \bibinfo{person}{Pierric Cistac}, \bibinfo{person}{Tim Rault},
  \bibinfo{person}{R'emi Louf}, \bibinfo{person}{Morgan Funtowicz}, {and}
  \bibinfo{person}{Jamie Brew}.} \bibinfo{year}{2019}\natexlab{}.
\newblock \showarticletitle{HuggingFace's Transformers: State-of-the-art
  Natural Language Processing}.
\newblock \bibinfo{journal}{{\em ArXiv\/}}  \bibinfo{volume}{abs/1910.03771}
  (\bibinfo{year}{2019}).
\newblock


\bibitem[\protect\citeauthoryear{Zhou, Lapedriza, Xiao, Torralba, and
  Oliva}{Zhou et~al\mbox{.}}{2014}]%
        {zhou2014learning}
\bibfield{author}{\bibinfo{person}{Bolei Zhou}, \bibinfo{person}{Agata
  Lapedriza}, \bibinfo{person}{Jianxiong Xiao}, \bibinfo{person}{Antonio
  Torralba}, {and} \bibinfo{person}{Aude Oliva}.}
  \bibinfo{year}{2014}\natexlab{}.
\newblock \showarticletitle{Learning deep features for scene recognition using
  places database}. In \bibinfo{booktitle}{{\em Advances in neural information
  processing systems}}. \bibinfo{pages}{487--495}.
\newblock


\end{thebibliography}

\end{document}